\pdfoutput=1
\documentclass[letterpaper]{article} % DO NOT CHANGE THIS
\usepackage{times}  % DO NOT CHANGE THIS
\usepackage{helvet}  % DO NOT CHANGE THIS
\usepackage{courier}  % DO NOT CHANGE THIS
\usepackage[hyphens]{url}  % DO NOT CHANGE THIS
\usepackage{graphicx} % DO NOT CHANGE THIS
\usepackage[numbers]{natbib}  % DO NOT CHANGE THIS AND DO NOT ADD ANY OPTIONS TO IT
\usepackage{caption} % DO NOT CHANGE THIS AND DO NOT ADD ANY OPTIONS TO IT
\usepackage{algorithm}
\usepackage{algorithmic}

\usepackage{newfloat}
\usepackage{listings}
\DeclareCaptionStyle{ruled}{labelfont=normalfont,labelsep=colon,strut=off} % DO NOT CHANGE THIS
\floatstyle{ruled}
\newfloat{listing}{tb}{lst}{}
\floatname{listing}{Listing}
\title{Neural Markov Prolog}
\author {
    Alexander Thomson,\textsuperscript{\rm 1}
    David Page \textsuperscript{\rm 2}
}
\date {
    \textsuperscript{\rm 1} Duke University\\
    \textsuperscript{\rm 2} Duke University\\[2ex]%
    \today 
}

\begin{document}

\maketitle

\begin{abstract}
%Modern implementations of deep neural network architectures tend to design around the fundamental assumptions of those models on a case by case basis that can make updating and designing new neural network structures based on new assumptions, difficult. In this paper, we propose the language Neural Markov Prolog (NMP) that is based in both Markov logic and Prolog as a means with which to bridge between first order logic and neural network designs and allow for the easy generation of specialized models.
The recent rapid advance of AI has been driven largely by innovations in neural network architectures.  A concomitant concern is how to understand these resulting systems. 
In this paper, we propose a tool to assist in both the design of further innovative architectures and the simple yet precise communication of their structure.  We propose the language Neural Markov Prolog (NMP), based on both Markov logic and Prolog, as a means to both bridge first order logic and neural network design and to allow for the easy generation and presentation of architectures for images, text, relational databases, or other target data types or their mixtures.
\end{abstract}

\section{Introduction}

%Neural network performance has made great strides in recent years by incorporating key assumptions, often referred to as inductive biases, on the domains of those specific tasks into the specialized structure of the model.
Neural network performance has made great strides in recent years by incorporating key assumptions, often referred to as inductive biases, about data domains into specialized model structures.
The designs of popular neural network architectures such as recurrent neural networks, convolutional neural networks, graph neural networks, and transformers all incorporate aspects of their respective task-specific domains into the operations, weight sharing, and connections of their underlying network structure \cite{dnnreview, recurrentnet, convolutionnet, graphnet, attentionnet}. That specialization, has, in turn, yielded improved efficiency and performance over the more general, fully connected design. Note, however, when implemented, these neural networks tend to be treated as entirely separate architectures, with no explicit connections between them, despite their similar underlying assumptions. Not only does this practice obscures some of the core theoretical similarities between these models, but it can also make modifying the architecture cumbersome when any of those original assumptions about the task domain change even slightly.

There exist several well-established methods for describing and reasoning from logical knowledge bases that could trivially describe both the assumptions made on a task’s domain and the graphical structure of the neural network itself. Nonetheless, simply using deterministic logic on its own to define that structure, through any given logical programming language, does not immediately align with the constrained structure of the neural network and the uncertainty present in said network’s predictions. On the other hand, methods from statistical relational learning that combine first order logic and probability can generate large and unwieldy probabilistic graphical models that lack the efficient training found in neural networks. In this paper, we therefore propose a new language with a basis in the languages of Prolog \cite{prologcite} and Markov logic \cite{richardson2006markov} can benefit from the expressivity of first order logic to describe the underlying assumptions on a task’s domain while simultaneously remaining grounded in the efficient training inherent to the neural network design. 

%With the recent proof in Li et al. \shortcite{LiThomsonEngelhardPage2023} that deep neural networks with sigmoid activations can be viewed as pairwise binary Markov networks where stochastic gradient descent in the neural network is either exactly or approximately equivalent to inference in the Markov network, depending on whether the network is expanded into an infinite tree structure, we show that, when viewed as Markov networks, deep neural networks can be expressed exactly in Markov logic given the language is sufficiently restricted. We then further limit this representation such that the gradient in the deep neural network now approximates, by way of standard stochastic gradient descent, the gradient of the Markov network defined in Markov logic so as to allow for efficient training. On its own, this limited form of Markov logic, in which each formula only involves a conjunction of two literal lays the groundwork for describing both the pairwise structure and parameter sharing of neural network designs in terms that are meaningful in first order logic. This is, in turn, the most basic form of the language proposed in the paper, which we name Neural Markov Prolog (NMP). By combining the logical foundation of both Prolog and Markov logic with the efficiency of modern neural networks, Neural Markov Prolog then not only provides a flexible framework with which to elegantly express pre-existing architectures, but also a tool that can be used to develop and present innovative new deep neural network designs in a meaningful and systematic manner.

We first show that all neural networks with only sigmoid activations can be
expressed in a highly-readable subset of Markov logic that is restricted to only binary (two-literal) clauses. 
This result follows formally from a proof in \cite{LiThomsonEngelhardPage2023} that all deep neural networks with only sigmoid activations can be viewed as binary pairwise Markov networks.
We then show how that restricted form of Markov logic can be extended simply
and elegantly to represent {\em all} neural networks in a syntactic form extremely close to the language Prolog, hence the name ``Neural Markov Prolog (NMP).''
By combining the logical foundation of both Prolog and Markov logic with the efficiency of modern neural networks, Neural Markov Prolog provides a flexible framework with which to elegantly express existing architectures, lending evidence to our claim that NMP may also be a useful tool to develop and present innovative new deep neural network designs in a meaningful and systematic manner.

NMP fits into the active research area of neuro-symbolic computation.  Nevertheless, most neuro-symbolic representations treat neural networks as black-box components of a larger symbolic knowledge representation \cite{SRLNeuroSymbol, neuralproblogicprogram, manhaeve2018deepproblog}. % Add at least a review of neuro-symbolic methods in addition to DeepProbLog here
In contrast, NMP brings explicit probabilistic semantics to the entire neuro-symbolic representation, and as such, it may be seen as unifying neuro-symbolic computation with statistical relational learning (SRL), where the meaning of any representation is itself a complete joint probability distribution with no black-box components.

\section{Deep Neural Networks as Markov Logic}

The equivalence between sigmoid neural networks and Markov logic networks emerges when the deep neural network itself is viewed as a Markov network. Recently Li et al. \cite{LiThomsonEngelhardPage2023} showed that any sigmoid neural network can be viewed as a binary pairwise Markov network, most easily seen using its loglinear formulation: 

$$P(\vec{v}) = \frac{1}{Z} \exp^{\sum_{i,j} w_{i,j} f_{i,j}(\vec{v})}$$ 

\noindent where each binary feature $f_{i,j}$ is 1 if and only if both nodes $i$ and $j$ are 1.
The nodes of this Markov network are the same as the nodes of the neural network, the features of this Markov network are the same as the edges of the neural network, and the weight on a feature of the Markov network is the same as the weight on the corresponding edge of the neural network.

Stochastic gradient descent (SGD) in the neural network is unfortunately not identical to exact probabilistic inference in the above Markov network and is instead an {\em approximation}. The Markov network has well-defined probabilistic semantics in the form of a joint probability distribution over all of its variables. Those precise probabilistic semantics can then be extended to the DNN context by characterizing the approximation SGD makes in the neural network with a Markov network.
Specifically, the approximation corresponds exactly to a simple, highly-regular but infinite, tree structured Markov network \cite{LiThomsonEngelhardPage2023}.
While one would never actually write down this infinite model, the theoretical contribution of the present paper is to show that this infinite Markov network is represented precisely by a finite Markov logic network (MLN), which therefore shares its semantics and which is both very easily written and understood.
Before presenting this specific MLN, we first review MLNs.

Richardson and Domingos \cite{richardson2006markov} define an MLN as a set of formula-weight pairs $(F_i, w_i)$ that, when grounded with a finite set of constants $C = \{c_1, …, c_{|C|}\}$, forms a Markov network with nodes for every ground predicate and edges drawn according to the cliques defined by each ground formula. Those cliques then have features $f_i$ and weights $w_i$ in the loglinear form of the network, where $f_i$ is 1 if and only if the corresponding ground formula is true and 0 otherwise. Just as in the Markov network view of deep neural networks, these features are binary in Markov logic. Clearly, not every MLN is equivalent to a DNN: if a formula contains more than two ground predicates, then the resulting network will lack the pair-wise connected structure inherent to the DNN design.

Nevertheless, in the following section we show that any sigmoid DNN is equivalent to a simple MLN.  Specifically, if the MLN has only formulas that involve the conjunction of two literals each with their own weight $w_i$ in the Markov logic network, the resulting Markov network will indeed have both the pairwise connected structure and binary features of the aforementioned DNN structure.
We later extend this MLN representation to allow other activation functions beyond sigmoid.

These restrictions to Markov logic do sacrifice its ability to concisely and clearly reason from {\em any} given logical knowledge base. Nonetheless, the resulting restricted language, which we name Neural Markov Prolog, can now benefit from the much more efficient training of deep neural networks while retaining a connection to first-order logic through its basis in Markov logic. As shown in our examples, that logical foundation can be used to elegantly express and explore a wide range of popular neural network architectures whose similar underlying assumptions become readily apparent when viewed through the lens of Neural Markov Prolog. 

\section{Neural Markov Prolog}

A Neural Markov Prolog program is built from two main components: an \textit{interpreted} section that involves the aforementioned pairs of literals and corresponding real-valued weights and a \textit{deterministic} section whose formulas are unrestricted and, in the Markov logic view of the NMP program, are each assigned an infinite weight. The deterministic piece of a Neural Markov Prolog program is used to describe known logical structure in the data and problem of interest, time-steps, neighboring pixels, etc, that might be relevant in the construction of the DNN. The interpreted section of the NMP program is then used to define, by referencing and leveraging that deterministic logical information, the pairwise structure and parameter sharing of the final neural network. 

Given that the deterministic section of the NMP program simply involves strict logical rules and facts, it can be evaluated on its own with any logical programming language. Prolog is used in the NMP program presented in this work and so this deterministic section’s syntax and semantics exactly matches Prolog’s syntax and semantics \cite{prologcite}. The interpreted portion of a Neural Markov Prolog (NMP) program similarly adopts a syntax in-line with pure Prolog with the notable differences in that each interpreted line has the form of a conjunction of two literals and a directionality that is expressed with the `:-` symbol as per a Prolog rule instead of the true underlying conjunction. Each line is also assigned a weight as shown in Figure \ref{fig1}. Note, however in the final version of NMP presented in this work, this weight is typically not written explicitly by the programmer. 

The interpreted NMP program takes its semantics from Markov logic when it creates the neural network's graphical structure. The resulting network can then also be thought as a probability distribution over the program's Herbrand Base, i.e. the set of ground atoms that can be generated from its vocabulary, in the manner described in the remainder of this section. We assume the reader is already familiar with Prolog and Markov logic. 

As with Markov logic, we go from an NMP interpreted program to the joint probability distribution it represents by first constructing an undirected graph whose nodes correspond to the ground atoms, which are the random variables in the probabilistic graphical model view. Also, as in Markov logic, an edge is placed between two nodes if and only if those nodes appear in any ground instance of some expression. Since every expression (or “rule”) in the interpreted NMP program is a conjunction of two literals, each of these ground expressions then correspond to a neuron pair in the neural network whose directionality is denoted with the same syntax as Prolog’s Horn clauses. To complete the construction of the Markov network that precisely defines the probability distribution, we take the weights on the cliques in the resulting graph from the weights on the expressions (rules in NMP).

By then referencing elements of the deterministic section of the NMP program, the interpreted section can integrate established domain knowledge and logical structure into the connections and weight sharing of this network. In the Markov logic view of the complete program as shown in Figure \ref{fig1}, this step simply involves the interpreted formulas referencing literals present in the infinite-weighted formulas of the deterministic section. In the final version of NMP presented in this work, we more clearly delineate between the deterministic and interpreted sections of the NMP program by including a Prolog \textit{query} term in each interpreted line that references the deterministic logic of the program and defines the possible groundings of that line's variables.

\section{A Simple Example}

\begin{figure}[t]
\centering
\includegraphics[width=0.8\columnwidth]{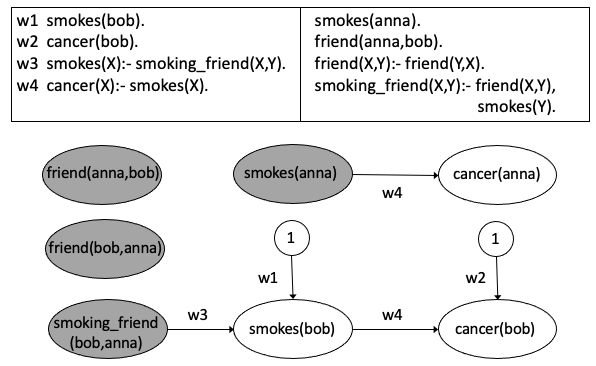}
\caption{A simple program of Neural Markov Prolog written in a style similar to Markov logic. Note that a missing weight before a clause indicates an infinite weight, or a traditional Prolog rule.}
\label{fig1}
\end{figure}

Consider, then, a simple database that records the friendships between individuals in a group and whether or not some of those individuals smoke. Suppose also there is missing information on the smoking status of several of the individuals recorded. Additionally, a rule is defined for whether an individual in the database has a friend who smokes. As shown in Figure \ref{fig1}, it is trivial to represent this information in Prolog. 

The interpreted piece of the Neural Markov Prolog program can then reference that stored information to incorporate those relationships fundamentally into the connections and parameter sharing of the neural network it defines. The interpreted section of the NMP program in Figure \ref{fig1} defines formulas for the relationship between smoking and having a friend who smokes and smoking and developing cancer. 

These formulas are then ground with the constants defined in the deterministic section of the program. For example, the variables $X$, $Y$ in 'smokes(X):-smoking\_friend(X,Y)' are ground to $bob$ and $anna$ respectively since 'smoking\_friend(bob,anna)' is the only valid deterministic grounding. On the other hand, the variable in $X$ in the formula 'cancer(X):-smokes(X)' could be either $bob$ or $anna$, which, in turn, results in multiple neuron connections all with a shared weight $w4$ in the DNN defined by the program. 

Two additional singular propositions (non-conjunctive) are also included in this example in order to define the biases of the resulting neurons in the DNN. Note, however, that while these term are necessary to define the biases in the original Markov logic definition of NMP, unless they are to be initialized with a specific value, they can be incorporated automatically into every neuron and as such are omitted from the program's text in the final version of the language. Similarly, nodes with no connection to any potential output node, neurons that lack any children in the most general case, or neuron simply not specified among the outputs of a provided data set, can also be ignored in the graphical structure defined by the NMP program. 

\section{The Exact Equivalence}

\begin{figure}[t]
\centering
\includegraphics[width=0.9\columnwidth]{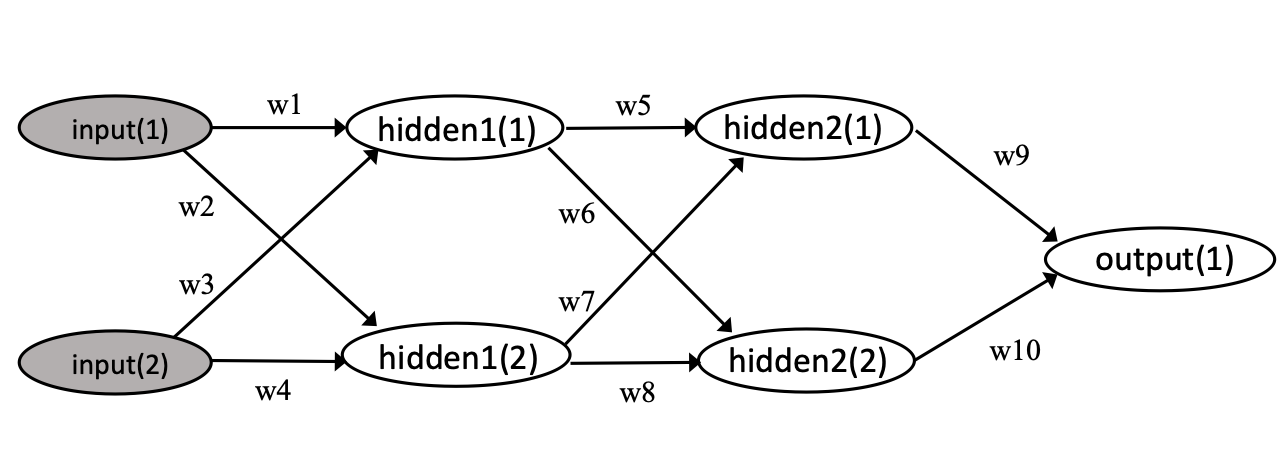}
\caption{The DNN structure generated by DNN.nmp.}
\label{fig2}
\end{figure}

\begin{figure}[t]
\centering
\includegraphics[width=0.9\columnwidth]{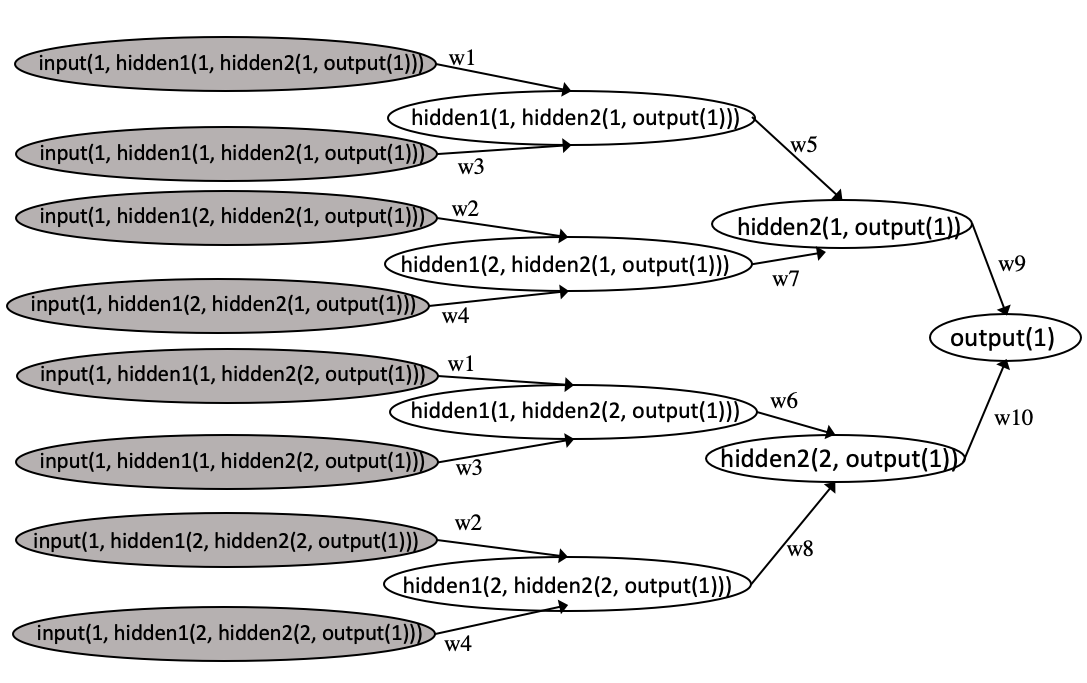}
\caption{Step 1 in the PGM construction. Parents nodes in the network are copied in order to construct a tree.}
\label{fig3}
\end{figure}

\begin{figure*}[t]
\centering
\includegraphics[width=1\columnwidth]{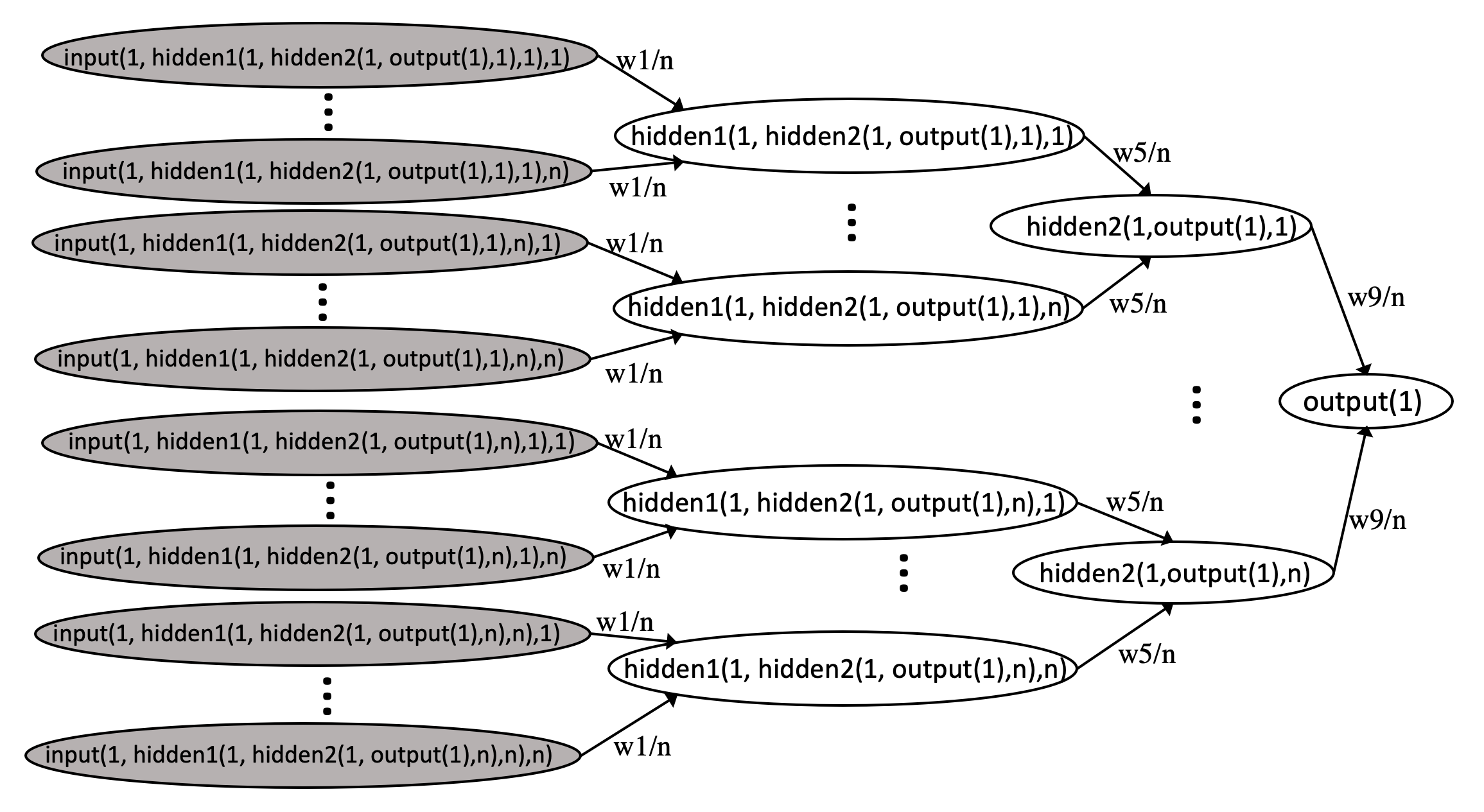}
\caption{Step 2 in the PGM construction applied to the upper nodes (weights w1, w5, w9) of the graph created in Figure \ref{fig3}. Each node is copied $n$ times along with its ancestors where $n$ is a large integer. Note that the ellipses in this figure are in the place of copied nodes, all of which would also have the same copied tree structure tying back to the inputs. As $n$ approaches infinity, this construction exactly characterizes the approximation made by SGD in the DNN.}
\label{fig4}
\end{figure*}

It is worth, noting, however, that the construction described so far will not necessarily generate a network in which stochastic gradient descent for the deep neural network exactly matches probabilistic inference in the Markov network.  Li, Thomson, Engelhard, and Page \cite{LiThomsonEngelhardPage2023} showed that the gradients of cross-entropy error used in training neural networks agree with the gradients of infinite tree-structured Markov networks. As such, the semantics or joint probability distribution represented by a neural network can be understood as represented by that specific Markov network structure.

Step 1 of this Markov network construction (Figure \ref{fig3}) gives every node $N$ its own unique copy of each parent $P$, and indeed of the entire network from the input nodes to $P$.
Step 2 of the construction then creates $L$ copies (where $L$ is a large integer and in the limit, $L \to \infty$) of each Step 1 copy along with copies of their ancestral nodes and edges, and divides the original weight $W$ from $N$ to $P$ by $L$; thus the weight of the edge into $N$ from any copy of $P$ is $\frac{W}{L}$. 

Each edge of the neural network generated by an NMP program comes from some specific clause with the general form: 
\begin{displaymath}
p(t_1, ..., t_n) \textsc{ :- } q(s_1, ..., s_m)
\end{displaymath}
Step 1 of the previous paragraph can be encoded by adding to each body literal in the NMP program one additional argument: the head literal, where its predicate is treated as a new function symbol within the body literal's predicate. Note that while the current definition of NMP does not, in fact, include functions terms that can be used to recursively reference the nodes used to define the structure of the deep neural network as is shown here, recording the graphical structure of defined DNN in the deterministic section of the program and then generating the network accordingly would yield the same result. For the sake of brevity, we present this form.
\footnote{Here the role of ``p'' as either predicate or function symbol is clear from the context, but if we wish to more explicitly distinguish them then we can make a consistent alteration to function symbols as here: $p(t_1, ..., t_n) \textsc{ :- } q(s_1, ..., s_m, p'(t_1, ..., t_n))$.}
\begin{displaymath}
p(t_1, ..., t_n) \textsc{ :- } q(s_1, ..., s_m, p(t_1, ..., t_n))
\end{displaymath}
Step 2 can be encoded by adding to the body literal of a clause yet one more argument, which is a list that appends a copy number for each instance of the body literal of a clause to the corresponding list from the head literal of the clause.
If the head literal has no such list (it has not been copied), then the list in the body literal has only a single element.
Steps 1 and 2 are illustrated in Figures \ref{fig3} and \ref{fig4} respectively, using the small, fully-connected network shown in Figure \ref{fig2}. The above discussion defines the semantics for pure Neural Markov Prolog. This is easily extended to use all of Prolog.

\section{The Finalized Syntax}

While the restricted form of Markov logic described up to this point would be able to define any neural network with sigmoid activations, for neural networks with a large number of unique weights, the number of formula-weight pairs needed to express even the most basic DNN designs would make this program needlessly complex. For that reason, our definition of Neural Markov Prolog also includes variables that can be marked as untethered. When an NMP program is evaluated, the unique combinations of these ground untethered variables are then used as what are effectively separate Markov logic formulas with their own unique weights. This addition allows for the concise definition of many formula-weight pairs while not comprising the theoretical background of the language as any NMP program written using these variables could also be written without them. 

Of course, even with these additions, certain popular features of modern neural networks such as dropout, batch normalization, or non-sigmoid activations have not been fully aligned theoretically with Markov logic. We therefore further expand upon the syntax and semantics of this foundational restricted Markov logic and add a final \textit{options} section to each interpreted NMP line so that neural networks that rely on these features still fall within the scope of what Neural Markov Prolog can express. With these two additions to the language, the interpreted lines in a NMP program then have the following form:

\begin{algorithmic}
\STATE head\_predicate(X1,...,Xn) :- body\_predicate(Y1,...,Ym),\newline
    \hspace*{11.5em}PROLOG QUERY, \newline 
    \hspace*{11.5em}+[OPTIONS].
\end{algorithmic}

\noindent In this version, the predicates in the body and head of each line are not automatically matched to the deterministic section of the program and, instead, the query term is used to ground each line in the NMP program. With this choice, the previous NMP code shown in Figure \ref{fig1} would be written as the following in order to generate the same neural network.

\begin{algorithm}
\caption*{\textbf{Smoking.nmp}}
    \begin{algorithmic}[1]
        \STATE participant(anna). 
        \STATE participant(bob).
        \STATE smokes(anna).
        \STATE friend(anna,bob).
        \STATE friend(X,Y) :- friend(Y,X).
    \end{algorithmic}
\end{algorithm}

\begin{algorithm}
\caption*{\textbf{Smoking.nmp}}
    \begin{algorithmic}[1]
        \STATE smokes(X) :- smoking\_friend(X,Y), \newline
                        \hspace*{5.2em} friend(X,Y), smokes(Y).
        \STATE cancer(X) :- smokes(X), participant(X).
    \end{algorithmic}
\end{algorithm}

\noindent Note that the terms used to define the biases on each neuron and the weights have both been removed and are now implicit in the lines of the NMP program.

Using this version of Neural Markov Prolog, it is then straightforward to define several fully connected deep neural network layers. In fact, due to the absence of strong assumptions made by this architecture on its data, the deterministic section of NMP program can remain entirely empty.

\subsection{A Fully Connected DNN}

\begin{algorithm}
\caption*{\textbf{DNN.nmp}}
    \begin{algorithmic}[1]
        \STATE hidden1(X) :- input(W), between(1,2,X), \newline
                    \hspace*{5.5em} between(1,2,W), +[untethered:X,W].
        \STATE hidden2(Y) :- hidden1(X), between(1,2,Y), \newline
                    \hspace*{5.5em} between(1,2,X), +[untethered:X,Y].
        \STATE output(1) :- hidden2(Y), between(1,2,Y), \newline
                    \hspace*{5em} +[untethered:Y]. 
    \end{algorithmic}
\end{algorithm}

\noindent The resulting deep neural network is shown in Figure \ref{fig2}. 

 Notably, these lines leverage the untethered variables option of the NMP program to concisely define this network in 3 lines. There are a total of 10 unique weights in the final DNN, which would otherwise require 10 unique formulas. Instead, these lines each define multiple weights as per the unique combinations of the ground untethered variables. This process can be thought of as expanding out each line to define multiple formulas each with its own unique weight as shown in the following code snippet (the unique weights are shown on the right).

\begin{algorithm}
\caption*{\textbf{DNN untethered parameters}}
    \begin{algorithmic}[1]
        \STATE hidden1(1) :- input(1), between(1,2,1), \newline
                    \hspace*{5.5em}between(1,2,1). \hspace*{5em} w1
        \STATE hidden1(2) :- input(1), between(1,2,2), \newline
                    \hspace*{5.5em}between(1,2,1). \hspace*{5em} w2
        \STATE hidden1(1) :- input(2), between(1,2,1), \newline
                    \hspace*{5.5em}between(1,2,2). \hspace*{5em} w3
        \STATE hidden1(2) :- input(2), between(1,2,2), \newline
                    \hspace*{5.5em}between(1,2,2). \hspace*{5em} w4
    \end{algorithmic}
\end{algorithm}

Since there are no other relevant variables to ground after that step, the program then simply uses each pair of literals to define the network. There are two unique neurons defined by the ground body literal and two neurons defined by the ground head literal all connected in the four possible pairwise combinations. Since each of these connections is implicitly assigned a unique weight, this NMP therefore defines a layer in the fully connected neural network. 

\section{Popular Architectures Expressed in NMP}

We then show how this simple language, when combined with a deterministic program written in Prolog can express several popular neural network architectures.

\subsection{The Recurrent Neural Network (RNN)}

Inherent to the recurrent neural network architecture is the assumption of an ordered sequential structure underlying the data. This aligns neatly with the basic functions of successor and predecessor in logic. In Neural Markov Prolog, the RNN architecture can then be naturally derived by simply describing this basic assumption in the deterministic piece of the program as shown in \textbf{RNN.pl} and \textbf{RNN.nmp}. In this example, we consider an underlying structure with three separate data entries recorded across 10 time steps. 

\begin{algorithm}
\caption*{\textbf{RNN.pl}}
    \begin{algorithmic}[1]
        \STATE timestep(T) :- between(1, 10, T).
        \STATE rnn\_in(X, Index) :- timestep(X), between(1,3,Index).
    \end{algorithmic}
\end{algorithm}

\begin{algorithm}
\caption*{\textbf{RNN.nmp}}
    \begin{algorithmic}[1]
        \STATE hidden(X) :- input(X, Index), \newline
                    \hspace*{5em} rnn\_in(X, Index), +[untethered: Index].
        \STATE hidden(Y) :- hidden(X), timestep(X),  \newline
                    \hspace*{5em} timestep(Y), Y is X+1.
    \end{algorithmic}
\end{algorithm}

\noindent This defines a simple RNN with ten hidden neurons each connected to three input neurons. The $untethered$ variable Index, that ranges between 1 and 3, defines the three separate input variables at any given time-step that are each connected to the hidden neuron with a unique weight. Expanding out these untethered variables yields the following weight structure:

\begin{algorithm}[H]
\caption*{\textbf{RNN untethered parameters}}
    \begin{algorithmic}[1]
        \STATE hidden(X) :- input(X, 1), rnn\_in(X,1) \hspace*{4em} w1
        \STATE hidden(X) :- input(X, 2), rnn\_in(X,2) \hspace*{4em} w2
        \STATE hidden(X) :- input(X, 3), rnn\_in(X,3) \hspace*{4em} w3
        \STATE hidden(X) :- hidden(Y), timestep(X),   \newline
                    \hspace*{5em} timestep(Y), Y is X+1. \hspace*{4.4em} w4
    \end{algorithmic}
\end{algorithm}

\noindent $X$ in the first NMP line is not specified as $untethered$ so the network's weights are shared across its possible groundings. Since $X$ is matched to the time-steps described in the NMP program's deterministic section, this in turn yields the weight sharing structure expected from the recurrent neural network as the weights between the inputs and hidden neurons of the network, $w1$, $w2$, $w3$, are then shared in the connections defined by those ten possible values of $X$.

The connections between hidden neurons is then captured by the second line in the NMP program where the variables $X$ and $Y$, which both refer to time-steps, are related through the expression 'Y is X+1'. Since the initial Prolog program defines a clear ordering in these time-steps, this query can then leverage that information to expand out the recursive structure of the RNN. Clearly if those hard logical assumptions are not in fact true in the underlying data, this network's structure will also be based around incorrect information, which in turn could hinder its performance.

\subsection{The Convolutional Neural Network (CNN)}

Much like the recurrent neural network, the convolutional neural network relies on a clear ordering of its underlying logical elements. In the Prolog section of the NMP program, this change in the structure of the data requires only the very simple addition of another dimension to define the CNN's grid with positional information, as shown in \textbf{CNN.pl} and \textbf{CNN.nmp}. The deterministic section of that program begins by defining a 10 by 10 grid of pixels expected from the input to the CNN along with a rule for whether a pixel is in a specific position relative to another. 

\begin{algorithm}
\caption*{\textbf{CNN.pl}}
    \begin{algorithmic}[1]
        \STATE pixel(X,Y) :- between(1,10,X), between(1,10,Y).
        \STATE relpos(A,B,C,D,X,Y) :- pixel(A,B), between(-1,1,X), \newline
                    \hspace*{9.5em}  between(-1,1,Y), C is A+X, \newline
                    \hspace*{9.5em}  D is B+Y, pixel(C,D).
    \end{algorithmic}
\end{algorithm}

\begin{algorithm}
\caption*{\textbf{CNN.nmp}}
    \begin{algorithmic}[1]
        \STATE hidden(A,B,K) :- input(C,D), \newline
                    \hspace*{7em} relpos(A,B,C,D,X,Y),  \newline
                    \hspace*{7em} between(1,1,K), \newline
                    \hspace*{7em} +[untethered: X,Y,K, \newline
                    \hspace*{8em} activation: relu].
    \end{algorithmic}
\end{algorithm}

\noindent In this example the specified $untethered$ variables $X$, $Y$, and $K$ correspond to the kernel height, width, and number of filters in the CNN. $K$ here is set to 1, but could easily range across far more values as is the case in most CNN architectures. Note that despite $X$ and $Y$ not appearing within either head or body predicate, since both variables are included in the untethered variables section and appear within the Prolog query, they are used to define the possible weights from the NMP line. This defines the nine weights of a 3 by 3 CNN kernel as shown in the following snippet.

\begin{algorithm}
\caption*{\textbf{CNN untethered parameters}}
    \begin{algorithmic}[1]
        \STATE hidden(A,B,1) :- input(C,D),  \newline
                    \hspace*{6.5em} relpos(A,B,C,D,-1,-1), \newline
                    \hspace*{6.5em} between(1,1,1) \hspace*{6em} w1
        \STATE hidden(A,B,1) :- input(C,D),  \newline
                    \hspace*{6.5em} relpos(A,B,C,D,0,-1), \newline
                    \hspace*{6.5em} between(1,1,1) \hspace*{6em} w2
        \STATE hidden(A,B,1) :- input(C,D),  \newline
                    \hspace*{6.5em} relpos(A,B,C,D,1,-1), \newline
                    \hspace*{6.5em} between(1,1,1) \hspace*{6em} w3
                    
        \item[\vdots]\begin{center}\end{center}

        \STATE hidden(A,B,1) :- input(C,D),  \newline
                    \hspace*{6.5em} relpos(A,B,C,D,1,1), \newline
                    \hspace*{6.5em} between(1,1,1) \hspace*{6em} w9
    \end{algorithmic}
\end{algorithm}

Since the variables $A$, $B$ and $C$, $D$ are not included within the untethered list, weights are shared across their groundings, and since they are tied to pixel values, as specified in the definition of the `relpos` rule in the Prolog section, this kernel is applied to all the pixels in the specified 10 by 10 input grid. 

This example also shows how the \textit{options} section can be used to specify other neural network features. Here `activation: relu` is specified so these neurons would use ReLU activations rather than the default sigmoid. Other specifications such as batch or layer normalization, drop-out, etc could be added in a similar manner. 

\subsection{The Graph Neural Network (GNN)}

To define the structure of a simple Graph Neural Network (GNN) in Neural Markov Prolog, we first define each underlying graph structure in the NMP program's deterministic section. A rule for whether these nodes are neighbors is also added. This is shown in \textbf{GNN.pl} and \textbf{GNN.nmp} where a small graph of four nodes $a$, $b$, $c$, and $d$ are connected in a simple diamond shape that is then used by the interpreted NMP program to create a neural network structure that passes information to the following layer based on those neighboring connections. 

\begin{algorithm}
\caption*{\textbf{GNN.pl}}
    \begin{algorithmic}[1]
        \STATE node(a). node(b). node(c). node(d).
        \STATE edge(a,b). edge(b,c). edge(c,d). edge(d,a).
        \STATE neighbor(X,Y) :- node(X), node(Y),  \newline
                    \hspace*{7em} (edge(X,Y) ; edge(Y,X)).
    \end{algorithmic}
\end{algorithm}

\begin{algorithm}
\caption*{\textbf{GNN.nmp}}
    \begin{algorithmic}[1]
        \STATE hidden(Y) :- input(X), node(X), node(Y), \newline
                    \hspace*{5em} (X == Y ; neighbor(X, Y)).
    \end{algorithmic}
\end{algorithm}

 \noindent Note that in the Neural Markov Prolog code, we do not specify any untethered variables, so this network only has a single weight that is shared across all ground pairs. Those pairs, in turn, are then between neighboring nodes and the node in the next layer, or the same node with itself between layers. It would, however, be entirely possible to incorporate multiple weights into this DNN structure by either creating NMP rules for other elements of the graph, such as the edges, or by simply including untethered variables into the head and body predicates of the existing rule. These choices depend heavily on the problem and data set of interest and as such necessitate a certain degree of flexibility in the GNN structure, which Neural Markov Prolog can easily provide.

\section{Conclusions and Future Work}

Neural Markov Prolog provides a flexible tool with which to both write and better interpret neural network designs. We have shown that NMP programs can concisely represent several popular neural network architectures in a process that involves first writing out the well-established assumptions on a problem of interest and then, from that foundational logic, establishing the full network's structure and weight sharing. Not only does this clear logical description of these foundational beliefs better convey the underlying structures of these models, but its inclusion also allows for the rapid development and testing of non-standard deep neural network architectures and provides the ability to update the fundamental assumptions that a DNN is based upon. We therefore claim that Neural Markov Prolog can be useful tool for the development and presentation of innovative new neural network designs.  Neural Markov Prolog enjoys Prolog's syntactic simplicity and power of concise representation, Markov logic's precise probabilistic semantics, and the computational efficiency and discovery power of deep neural networks.

There has been interest recently in automatic discovery of new neural network architectures, for example in the training of Google's LaMDA LLM \cite{thoppilan2022lamda}.  The logical representation of NMP means that pre-existing tools such as inductive logic programming and SRL could be directly applied to rule discovery in NMP, and hence to training new deep architectures.  This approach could in turn yield neural networks that are far more interpretable and perhaps even allow for the NMP network to expand upon itself.

One of the central future concerns for the development of Neural Markov Prolog is the optimization of the many operations in the neural network generated by the NMP program. While it is straightforward to define all pairs of neurons individually, that implementation would lack the efficiency of matrix operations often found in practical neural network implementations.  Our current NMP implementation directly compiles to matrix operations, but a more intelligent compiler could find more opportunities for matrix calculations.

Public databases and knowledge bases such as in biology (e.g., KEGG or Database of Interacting Proteins) can be directly compiled into definite clause logic, and therefore into NMP. Automatically translating some of the many relational databases for specific scientific data and problems into NMP programs could yield a more effective and efficient approach to neural network design for specialized tasks.

% \section{Acknowledgements}

\bibliography{nmp}

\begin{thebibliography}{10}

\bibitem{dnnreview}
L.~Alzubaidi, J.~Zhang, A.~Humaidi, A.~Al-Dujaili, Y.~Duan, O.~Al-Shamma,
  J.~Santamaría, M.~Fadhel, M.~Al-Amidie, and L.~Farhan.
\newblock Review of deep learning: concepts, cnn architectures, challenges,
  applications, future directions.
\newblock {\em Journal of Big Data}, 8, 03 2021.

\bibitem{prologcite}
A.~Colmerauer and P.~Roussel.
\newblock The birth of prolog.
\newblock {\em SIGPLAN Not.}, 28(3):37–52, mar 1993.

\bibitem{recurrentnet}
M.~I. Jordan.
\newblock Chapter 25 - serial order: A parallel distributed processing
  approach.
\newblock In J.~W. Donahoe and V.~{Packard Dorsel}, editors, {\em
  Neural-Network Models of Cognition}, volume 121 of {\em Advances in
  Psychology}, pages 471--495. North-Holland, 1997.

\bibitem{convolutionnet}
Y.~LeCun, B.~Boser, J.~Denker, D.~Henderson, R.~Howard, W.~Hubbard, and
  L.~Jackel.
\newblock Handwritten digit recognition with a back-propagation network.
\newblock In D.~Touretzky, editor, {\em Advances in Neural Information
  Processing Systems}, volume~2. Morgan-Kaufmann, 1989.

\bibitem{LiThomsonEngelhardPage2023}
B.~Li, A.~J. Thomson, M.~M. Engelhard, and D.~Page.
\newblock On neural networks as infinite tree-structured probabilistic
  graphical models, 2023.

\bibitem{manhaeve2018deepproblog}
R.~Manhaeve, S.~Dumancic, A.~Kimmig, T.~Demeester, and L.~De~Raedt.
\newblock Neural probabilistic logic programming in deepproblog.
\newblock {\em Artificial Intelligence}, 298:103504, 04 2021.

\bibitem{SRLNeuroSymbol}
L.~d. Raedt, S.~Dumančić, R.~Manhaeve, and G.~Marra.
\newblock From statistical relational to neuro-symbolic artificial
  intelligence.
\newblock In C.~Bessiere, editor, {\em Proceedings of the Twenty-Ninth
  International Joint Conference on Artificial Intelligence, {IJCAI-20}}, pages
  4943--4950. International Joint Conferences on Artificial Intelligence
  Organization, 7 2020.

\bibitem{richardson2006markov}
M.~Richardson and P.~Domingos.
\newblock Markov logic networks.
\newblock {\em Machine Learning}, 62:107--136, 02 2006.

\bibitem{graphnet}
F.~Scarselli, M.~Gori, A.~C. Tsoi, M.~Hagenbuchner, and G.~Monfardini.
\newblock The graph neural network model.
\newblock {\em IEEE Transactions on Neural Networks}, 20(1):61--80, 2009.

\bibitem{neuralproblogicprogram}
L.~D. Smet, P.~Z.~D. Martires, R.~Manhaeve, G.~Marra, A.~Kimmig, and L.~D.
  Raedt.
\newblock Neural probabilistic logic programming in discrete-continuous
  domains.
\newblock {\em CoRR}, abs/2303.04660, 2023.

\bibitem{thoppilan2022lamda}
R.~Thoppilan, D.~D. Freitas, J.~Hall, N.~Shazeer, A.~Kulshreshtha, H.-T. Cheng,
  A.~Jin, T.~Bos, L.~Baker, Y.~Du, Y.~Li, H.~Lee, H.~S. Zheng, A.~Ghafouri,
  M.~Menegali, Y.~Huang, M.~Krikun, D.~Lepikhin, J.~Qin, D.~Chen, Y.~Xu,
  Z.~Chen, A.~Roberts, M.~Bosma, V.~Zhao, Y.~Zhou, C.-C. Chang, I.~Krivokon,
  W.~Rusch, M.~Pickett, P.~Srinivasan, L.~Man, K.~Meier-Hellstern, M.~R.
  Morris, T.~Doshi, R.~D. Santos, T.~Duke, J.~Soraker, B.~Zevenbergen,
  V.~Prabhakaran, M.~Diaz, B.~Hutchinson, K.~Olson, A.~Molina, E.~Hoffman-John,
  J.~Lee, L.~Aroyo, R.~Rajakumar, A.~Butryna, M.~Lamm, V.~Kuzmina, J.~Fenton,
  A.~Cohen, R.~Bernstein, R.~Kurzweil, B.~Aguera-Arcas, C.~Cui, M.~Croak,
  E.~Chi, and Q.~Le.
\newblock Lamda: Language models for dialog applications, 2022.

\bibitem{attentionnet}
A.~Vaswani, N.~Shazeer, N.~Parmar, J.~Uszkoreit, L.~Jones, A.~N. Gomez, L.~u.
  Kaiser, and I.~Polosukhin.
\newblock Attention is all you need.
\newblock In I.~Guyon, U.~V. Luxburg, S.~Bengio, H.~Wallach, R.~Fergus,
  S.~Vishwanathan, and R.~Garnett, editors, {\em Advances in Neural Information
  Processing Systems}, volume~30. Curran Associates, Inc., 2017.

\end{thebibliography}

\end{document}